\documentclass[letterpaper, 10 pt, conference]{ieeeconf}  % Comment this line out if you need a4paper
\usepackage{times}
\usepackage{epsfig}
\usepackage{multirow}
\usepackage{xparse}
\usepackage[draft]{hyperref}
\usepackage{url}            % simple URL typesetting
\usepackage{booktabs}       % professional-quality tables
\usepackage{amsfonts}       % blackboard math symbols
\usepackage{nicefrac}       % compact symbols for 1/2, etc.
\usepackage{microtype}      % microtypography
\usepackage{cite}
\usepackage{epstopdf}
\usepackage{rotating,booktabs,multirow}
\usepackage{amsmath,amssymb} % define this before the line numbering.
\usepackage{color}
\usepackage{subcaption}
\usepackage{wrapfig,lipsum}
\usepackage{tabularx}
\usepackage{algorithm}
\usepackage{algorithmic}
\usepackage[algo2e]{algorithm2e}
\usepackage{mathtools}
\newcommand{\etal}{\textit{et al. }}
\newcommand{\minF}{\mathop{\mathrm{min}}\limits}
\newcommand{\maxF}{\mathop{\mathrm{max}}\limits}

\IEEEoverridecommandlockouts                              % This command is only needed if 
\title{\LARGE \bf Unsupervised Pixel-level Road Defect Detection\\via Adversarial Image-to-Frequency Transform}
\author{Jongmin Yu$^{1,2}$, Duyong Kim$^{3}$, Younkwan Lee$^{2}$, Moongu Jeon$^{2,\dagger}$% <-this % stops a space
\thanks{$^{1}$Curtin University, Kent St, Bently WA 6102, Australia {\tt\small$^{1,2}$jm.andrew.yu@gmailcom}}%
\thanks{$^{2}$Gwangju Institute of Science and Technology (GIST), Gwangju 61005, South Korea {\tt\small\{$^{2}$brightyoun, $^{2,\dagger}$mgjeon\}@gist.ac.kr}}%
\thanks{$^{3}$RMIT University, 124 La Trobe St, Melbourne VIC 3000, Australia {\tt\small$^{3}$duyong.kim@rmit.edu.au}}}

\begin{document}
\maketitle
\thispagestyle{empty}
\pagestyle{empty}

\begin{abstract}
In the past few years, the performance of road defect detection has been remarkably improved thanks to advancements on various studies on computer vision and deep learning. Although a large-scale and well-annotated datasets enhance the performance of detecting road defects to some extent, it is still challengeable to derive a model which can perform reliably for various road conditions in practice, because it is intractable to construct a dataset considering diverse road conditions and defect patterns. To end this, we propose an unsupervised approach to detecting road defects, using Adversarial Image-to-Frequency Transform (AIFT). AIFT adopts the unsupervised manner and adversarial learning in deriving the defect detection model, so AIFT does not need annotations for road defects. We evaluate the efficiency of AIFT using GAPs384 dataset, Cracktree200 dataset, CRACK500 dataset, and CFD dataset. The experimental results demonstrate that the proposed approach detects various road detects, and it outperforms existing state-of-the-art approaches.
\end{abstract}

\section{INTRODUCTION}
Road defect detection is one of the important studies to prevent vehicle accidents and manage the road condition effectively. All over the United States, road conditions contribute to the frequency and severity of motor vehicle accidents. Almost of third of all motor vehicle crashes are related to poor road conditions, resulting in more than two million injuries and 22,000 fatalities \cite{zaloshnja2009cost}. Over time, as road infrastructure ages, the condition of that infrastructure steadily declines, and the volumes and severity of defects increase \cite{carr2018road}. Therefore, the need for the development of a method for detecting road defects within this area only increases \cite{hadavandsiri2019concrete}, and numerous studies have been being proposed in the literature.  

Over the past decades, diverse studies have considered the use of image processing and machine learning approaches with hand-crafted features \cite{acosta1992low,bray2006neural,chambon2010road,deutschl2004defect}. Statistical analysis \cite{acosta1992low,chambon2010road} is the oldest one and also the most popular. Acosta \etal \cite{acosta1992low} and Deutschl \etal \cite{deutschl2004defect} have proposed vision-based methods based on partial differential techniques. Chambon \etal \cite{chambon2010road} have presented a method based on Markovian modelling to take into account the local geometrical constraints about road cracks. Bray \etal \cite{bray2006neural} have utilized the classification approach using neural networks for identifying road defects. These approaches usually identify road defects using the contrast of texture information on a road surface.

However, the contrast between roads and the defects on the roads may be reduced due to the illumination conditions and the changes in weather \cite{sun2009automated}. Additionally, the specification of cameras for capturing the surface of the roads also can affect the detection accuracies. Hense, it is still challenging to develop a defect detection method which can cover various road conditions in the real world using a simple image processing or machine learning methods alone \cite{baygin2015new}. 

Recently, various approaches \cite{pauly2017deeper,fan2019road} based on deep learning have been proposed to overcome these drawbacks. Pauly \etal \cite{pauly2017deeper} have proposed a method for road defect detection employing convolutional neural networks (CNNs). Fan \etal \cite{fan2019road} have proposed segmentation method based on CNNs and apply an adaptive. These approaches need a well-annotated dataset for road defects, and also their performance may depend on scale of the given dataset. Regrettably, it is problematic in practice to construct such a dataset containing various patterns of road defects.

Developing an unsupervised method which does not need annotations for road defects in the training step, is an issue that has been noticed for a long time in this literature. Various unsupervised approaches based on image processing and machine learning were proposed \cite{abdel2006pca,oliveira2012automatic}. However, these approaches still have an inherent weakness which is detection performances are highly dependent on camera specifications and image qualities. Recently, among the approaches based on deep learning, several studies \cite{mujeeb2019one,kang2018deep} have presented unsupervised methods using autoencoder \cite{vincent2010stacked}. These approaches take normal road images as their training samples and optimize their models in a way to minimize reconstruction errors between their input and output. These approaches recognize defects if the reconstruction errors of inputted samples are larger than a predefined threshold.

However, according to Perera \etal \cite{perera2019ocgan} and Pidhorskyi \etal \cite{pidhorskyi2018generative}, even though a model based on the reconstruction setting obtains a well-optimized solution, there is a possibility that the model can reconstruct samples which have not appeared in the training step. It could be a significant disadvantage in detecting road defects using the model. Due to this disadvantage, the model may produce lower error than the expectation even if it takes defect samples as their input, and it can make hard to distinguish whether this sample contains defects or not.

\begin{figure*}[ht]
	\vspace{-0.2cm}
	\centering
	\includegraphics[width=0.9\textwidth]{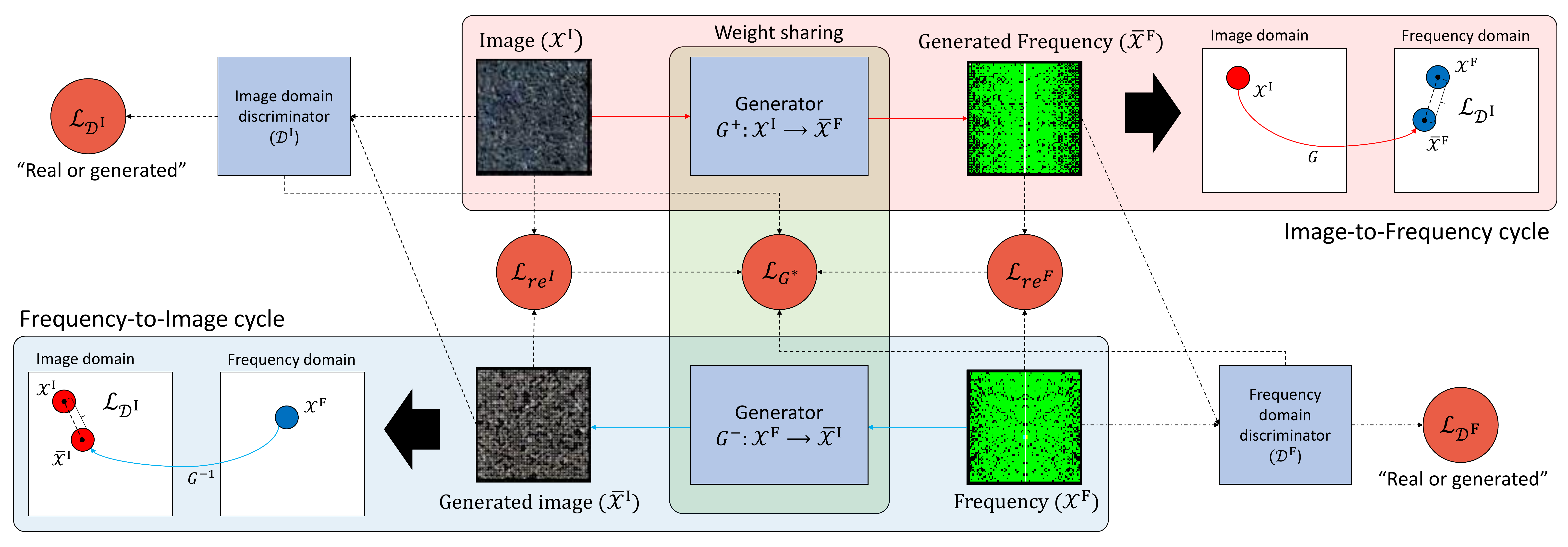}
	\caption{Architectural detail of the adversarial image-to-frequency transform. The blue objects denote the operation units including the generator $G$ and the discriminators $\mathcal{D}^{I}$ and $\mathcal{D}^{F}$. The red circles indicate the loss functions corresponded to the each operation unit. The red arrow lines show the work flow for the image-to-frequency cycle $G^{+}: \mathcal{X}^{I}\rightarrow{}\bar{\mathcal{X}}^{F}$, and the blue arrow lines represent the process of the frequency-to-image cycle $G^{-1}:\mathcal{X}^{F}\rightarrow{}\bar{\mathcal{X}}^{I}$. The dotted arrow lines represent the correlations of each component to the loss functions.}
	\label{iv_fig2}
	\vspace{-0.2cm}
\end{figure*}

To tackle this issue, we present an unsupervised approach, which exploits domain transformation based on adversarial learning, to detecting road defects. The proposed approach called Adversarial Image-to-Frequency Transform (AIFT) is trained by normal road images only and needs no annotations for defects. In contrast to other approaches \cite{mujeeb2019one,kang2018deep} optimizing their models by minimize reconstruction errors, AIFT is concentrated on deriving mapping function between an image-domain and a frequency-domain using adversarial manner. To demonstrate the efficiency of the proposed approach for road defect detection, we compare the proposed approach with various state-of-the-art approaches, including supervised and unsupervised methods. The experimental results show that the proposed approach can outperform existing state-of-the-art methods.

The main contributions of our work are summarized as follows:
\begin{itemize}
\item An unsupervised method for detecting road defects, which can provide outstanding performance without a well-annotated dataset for road defects.
\item The adversarial learning for deriving the image-to-frequency mapping function. Our approach can derive the more optimal transform model than typical approaches such as reconstruction or classification settings.
\item The extensive experiments about road defect detection. The experiments include ablation analysis depending on the loss functions and comprehensive comparison with the existing state-of-the-art methods.
\end{itemize}
In the further sections, we describe the details of our approach and provide the experimental results and analysis it. We conclude this paper by summarizing our works.

\section{THE PROPOSED METHOD}
\label{sec:3}
\subsection{Adversarial Image-to-Frequency Transform}
\label{sec:3:1}
It is essential to derive a robust model invariant to environments in order to detect a great number of defect patterns on roads. Our method is inspired by novelty detection studies \cite{perera2019ocgan,pidhorskyi2018generative}, which derive a model using inlier samples only and recognize outliers by computing a likelihood or an reconstruction error. The proposed method, called Adversarial Image-to-Frequency Transform (AIFT), initially derives a transform model between image-domain and frequency-domain using normal road pavement images only. The frequency-domain corresponding to the image-domain is generated by applying Fourier transform to the given image-domain. Detecting road defects is conducted by comparing given and generated samples of each domain. 

AIFT is composed of three components: Generator $G$, Image discriminator $\mathcal{D}^{I}$, Frequency discriminator $\mathcal{D}^{F}$, for applying adversarial learning.  The original intention of adversarial learning is to learn generative models while avoiding approximating many intractable probabilistic computations arising in other strategies \textit{e.g.,} maximum likelihood estimation. This intention is suitable to derive an optimal model for covering the various visual patterns of road defects. The workflow of AIFT is illustrated in Fig \ref{iv_fig2}. 

The generator $G$ plays as a role for the mapping function between image-domain $\boldsymbol{\mathcal{X}}^{I}=\{\mathcal{X}^{I}_{i}\}_{i=1:n}$ to frequency-domain $\boldsymbol{\mathcal{X}}^{F}=\{\mathcal{X}^{F}_{i}\}_{i=1:n}$ as follows, $G:\mathcal{X}^{I}\longleftrightarrow{}\mathcal{X}^{F}$. For the convenience of notation, we distinguish the notations of mappings for image-to-frequency $G^{+}:\boldsymbol{\mathcal{X}}^{I}\rightarrow{}\boldsymbol{\mathcal{X}}^{F}$ and frequency-to-image  $G^{-}:\boldsymbol{\mathcal{X}}^{F}\rightarrow{}\boldsymbol{\mathcal{X}}^{I}$, separately. $G$ generate the transformed results from each domain as follows,
\begin{equation}
\begin{split}
&G^{+}(\mathcal{X}^{I})=\bar{\mathcal{X}}^{F},\\
&G^{-}(\mathcal{X}^{F})=\bar{\mathcal{X}}^{I},
\end{split}
\end{equation}
where $\bar{\mathcal{X}}^{F}$ and $\bar{\mathcal{X}}^{I}$ indicate the transformed results from $\mathcal{X}^{I}$ and $\mathcal{X}^{F}$, respectively. $\bar{\mathcal{X}}^{I}$ and $\bar{\mathcal{X}}^{F}$ are conveyed to the two discriminators $\mathcal{D}^{I}$ and $\mathcal{D}^{F}$ for computing an adversarial loss. For computational-cost-effective implementation, weight sharing has employed.

\begin{figure}[t]
	\vspace{-0.2cm}
	\centering
	\includegraphics[width=\columnwidth]{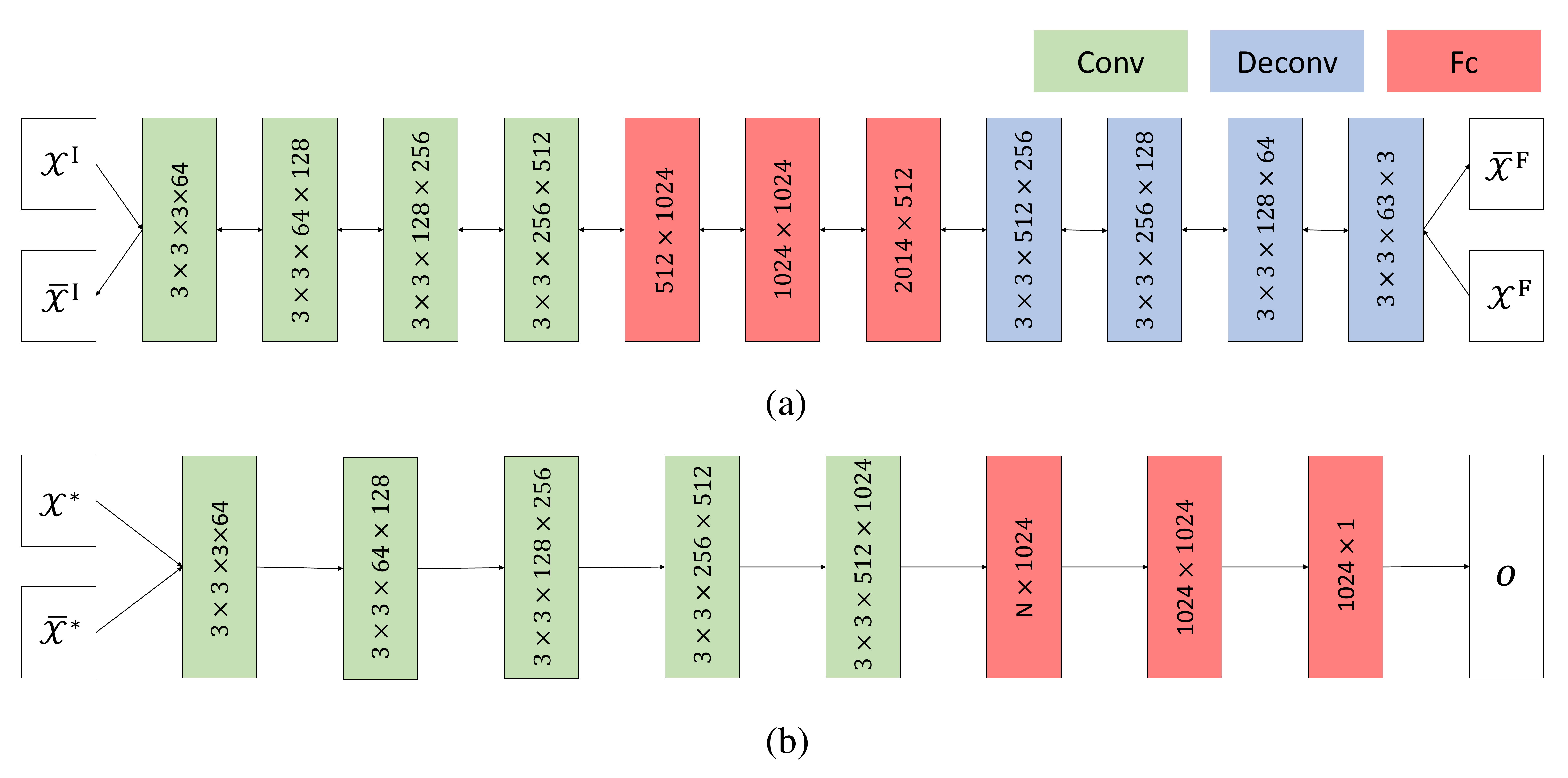}
	\caption{Structural details of the network models in the generator $G$ and the discriminators $\mathcal{D}^{I}$ and $\mathcal{D}^{F}$. (a) and (b) denote the structural details of the generator $G$ and the two discriminators $\mathcal{D}^{I}$ and $\mathcal{D}^{F}$, respectively. The green, blue, and red boxes denote the convolutional layers, the deconvolutional layers, and the fully-connected layers, respectively.}
	\label{iv_fig3}
	\vspace{-0.4cm}
\end{figure}

The discriminators $\mathcal{D}^{I}$ and $\mathcal{D}^{F}$ are defined as follows, 
\begin{equation}
\begin{split}
&\mathcal{D}^{*}(\mathcal{X}^{*})=o^{*},\quad{}o^{*}\in{}R^{1},
\end{split}
\end{equation}
where $*$ denotes the indicator to assign the discriminators $\mathcal{D}^{*}\in\{\mathcal{D}^{I},\mathcal{D}^{F}\}$ depending on the types of inputs $\mathcal{X}^{*}\in\{\mathcal{X}^{I},\mathcal{X}^{F},\bar{\mathcal{X}}^{I},\bar{\mathcal{X}}^{F}\}$. $\mathcal{D}^{I}$ takes $\mathcal{X}^{I}$ and $\bar{\mathcal{X}}^{I}$ as an input, and  $\mathcal{D}^{I}$ takes $\mathcal{X}^{F}$ and $\bar{\mathcal{X}}^{F}$ as an input, respectively. $o^{*}$ indicates the outputs $o^{I}$ and $o^{F}$ according to the types of the inputs and the discriminators. The value of $o^{*}$ can be regarded by as a likelihood to discriminate whether a given sample is truth or generated. Each component is compiled by CNNs and fully-connected neural networks and the structural details of these components are shown in Fig \ref{iv_fig3}.

\subsection{Adversarial transform consistency learning}
\label{sec:3:2}
As the workflow of AIFT shown in Fig \ref{iv_fig2}, the generator $G$ plays a role as a bidirectional mapping function between image-domain $\boldsymbol{\mathcal{X}}^{I}$ and corresponding frequency-domain $\boldsymbol{\mathcal{X}}^{F}$ generated from $\boldsymbol{\mathcal{X}}^{I}$. The underlying assumption for detecting road defects using AIFT is as follows. Since AIFT is only trained with normal road pavement images, if AIFT takes images containing defect patterns as an input, the error between the given samples and the transformed results would be larger than normal ones. Given this assumption, the prerequisite for precise road defect detection on AIFT is deriving a strict transform model between the image-domain and the frequency-domain from a given dataset for normal image samples for road pavement.

To end this, we present an adversarial transform consistency loss for training AIFT. Adversarial transform consistency loss is defined by,
\begin{equation}
\begin{split}
\mathcal{L}_{\text{ATCL}}(G,\mathcal{D}^{I},\mathcal{D}^{F})&=E_{\mathcal{X}^{I}\sim{}p_{\boldsymbol{\mathcal{X}^{I}}}}[\text{log}\mathcal{D}^{I}(\mathcal{X}^{I})]\\
&+E_{\mathcal{X}^{F}\sim{}p_{\boldsymbol{\mathcal{X}^{F}}}}[\text{log}\mathcal{D}^{F}(\mathcal{X}^{F})]\\
&+E_{\bar{\mathcal{X}}^{F}\sim{}p_{G^{^{+}}(\boldsymbol{\mathcal{X}^{I}})}}[\text{log}(1-\mathcal{D}^{F}(G^{+}(\mathcal{X}^{I})))]\\
&+E_{\bar{\mathcal{X}}^{I}\sim{}p_{G^{^{-}}(\boldsymbol{\mathcal{X}}^{F})}}[\text{log}(1-\mathcal{D}^{I}(G^{-}(\mathcal{X}^{F})))],
\end{split}
\label{eq:atcl_loss}
\end{equation}
where $G$ tries to generate images $\bar{\mathcal{X}}^{I}$ and frequency samples $\bar{\mathcal{X}}^{F}$ via $G^{+}$ and $G^{-}$ that look similar to
given images $\mathcal{X}^{I}$ and frequencies $\mathcal{X}^{F}$, while $\mathcal{D}^{I}$ and $\mathcal{D}^{F}$ aim to distinguish between given samples ($\mathcal{X}^{I}$ and $\mathcal{X}^{F}$) and transformed results ($\bar{\mathcal{X}}^{I}$ and $\bar{\mathcal{X}}^{F}$).

\begin{figure}[t]
	\vspace{-0.2cm}
	\centering
	\includegraphics[width=\columnwidth,height=4cm]{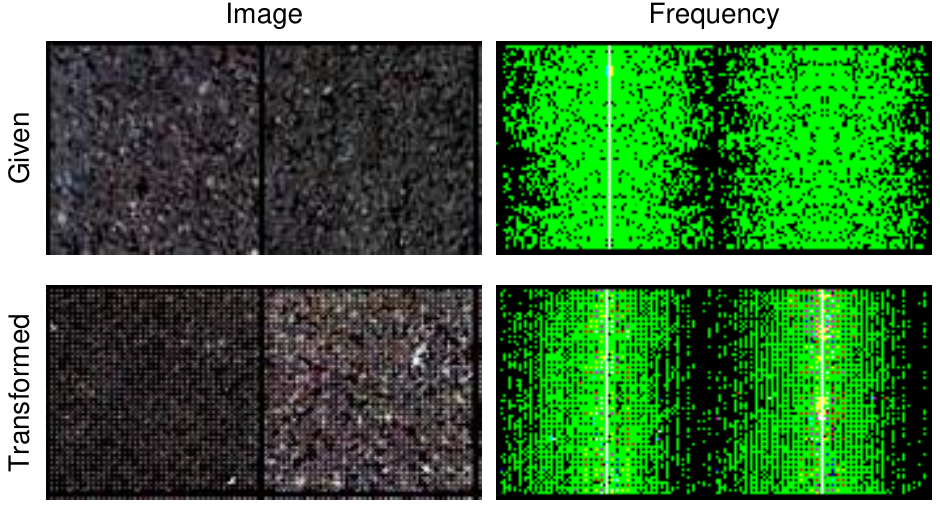}
	\caption{Comparison of the given and generated samples for the road pavement image and the corresponding frequency.}
	\label{iv_fig4}
	\vspace{-0.4cm}
\end{figure}

Adversarial learning can, in theory, learn mappings $G$ that produce outputs identically distributed as image and frequency domains, respectively \cite{zhu2017unpaired}. However, with large enough capacity, $G$ can map the same samples of an input domain to any random permutation of samples in the different domain, where any of the learned mappings can induce an output distribution that matches the target distribution. Thus, adversarial transform consistency loss alone may not guarantee that the learned function can map an individual input to the desired output.

To further reduce the space of possible mapping functions, we utilize the reconstruction loss to optimize the generator $\mathcal{G}$. It is a common way to enforce the output of the generator to be close to the target through the minimization of the reconstruction error based on the pixel-wise mean square error (MSE) \cite{ying2019x2ct,bai2018finding,liu2018future,sabokrou2018deep}. It is calculated in the form
\begin{equation}
    \begin{split}
\mathcal{L}_{\text{re}}(G) &= \mathbb{E}_{\mathcal{X}^{I}\sim{}p_{\boldsymbol{\mathcal{X}^{I}}}}[\Vert\mathcal{X}^{F}-G^{+}(\mathcal{X}^{I})\Vert^{2}_{2}]\\
&+E_{\mathcal{X}^{F}\sim{}p_{\boldsymbol{\mathcal{X}^{F}}}}[\Vert\mathcal{X}^{I}-G^{-}(\mathcal{X}^{F})\Vert^{2}_{2}].
    \label{eq:recon_loss}
    \end{split}
\end{equation}

Consequently, the total loss function is:, 
\begin{equation}
    \begin{split}
\mathcal{L}_{\text{total}}(G,\mathcal{D}^{I},\mathcal{D}^{F}) = \mathcal{L}_{\text{ATCL}}(G,\mathcal{D}^{I},\mathcal{D}^{F})+\lambda\mathcal{L}_{\text{re}}(G)
    \label{eq:total_loss}
    \end{split}
\end{equation}
where $\lambda$ indicates the balancing parameter to take the weight for the reconstruction loss.

Given the definition of above loss functions, the discriminators and the generator are trained by maximizing or minimizing corresponding loss terms expressed by,
\begin{equation}
\begin{split}
\text{arg}\minF_{\theta^{G}}\maxF_{\theta^{I},\theta^{F}}\mathcal{L}_{\text{total}}(G,\mathcal{D}^{I},\mathcal{D}^{F}),
\end{split}
\end{equation}
where $\theta^{G}$, $\theta^{I}$,and $\theta^{F}$ denote the parameters corresponded to the generator $\mathcal{G}$, the image discriminators $\mathcal{D}^{I}$, and the frequency discriminator $\mathcal{D}^{F}$. Fig \ref{iv_fig4} illustrates the examples of the given samples and the transformed results for image and frequency domains. We have conducted the ablation studies to observe the effect of each loss term in learning AIFT. 

\begin{figure*}[th]
    \centering
    \begin{subfigure}{0.47\textwidth}
        \includegraphics[width=\textwidth,height=3cm]{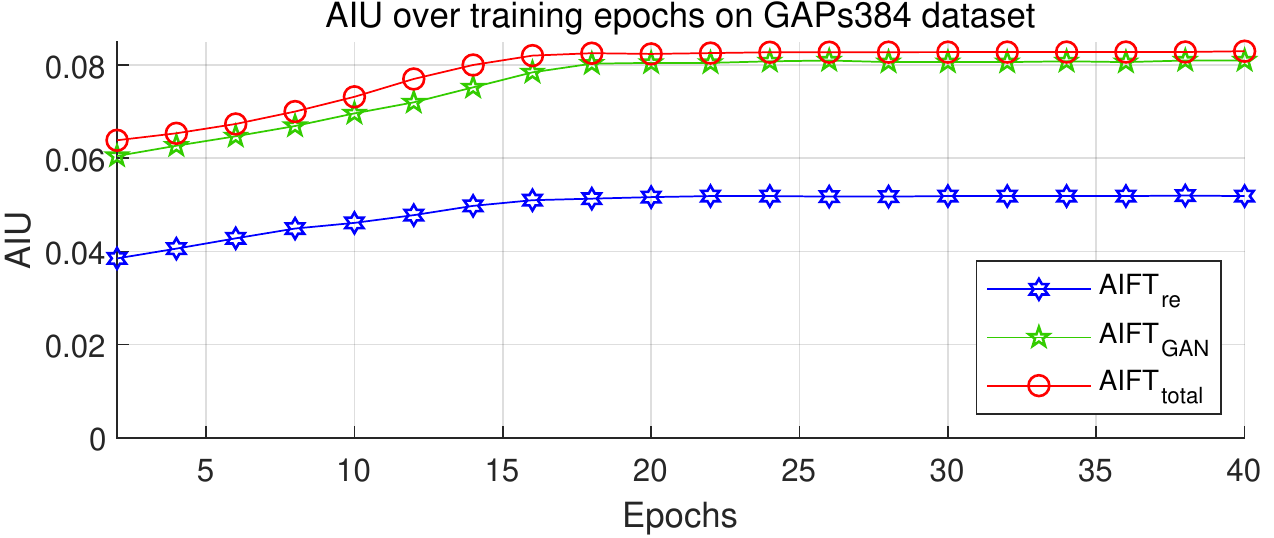}
    \caption{}
    \end{subfigure}
    \hfill
    \begin{subfigure}{0.47\textwidth}
        \includegraphics[width=\textwidth,height=3cm]{./figures/iv20_graph1.pdf}
    \caption{} 
    \end{subfigure}
    \hfill
    %\vspace{-1ex}
    \caption{The trends of AIU over the training epochs. (a) show the AIU trend over the training epochs on GAPs384 dataset, and (b) illustrate the AIU trend with respect to the training epochs on CFD dataset. The red-coloured curve (AIFT$_{\text{total}}$) denotes the AIU trend of AIFN trained by the total loss (Eq \ref{eq:total_loss}). The green-colored curve (AIFT$_{\text{GAN}}$) indicates the AIU trend of AIFN trained by the ATCL loss (Eq \ref{eq:atcl_loss}) only. The blue-colored curve (AIFT$_{\text{re}}$) shows the AIU trend of AIF trained by the reconstruction loss (Eq \ref{eq:recon_loss}).}
\label{fig:ab_study}   
\vspace{-2ex}
\end{figure*}

\subsection{Road defect detection}
Detecting defects on a road is straightforward. Initially, AIFT produces the frequency sample $\mathcal{X}^{F}$ using given an image samples $\mathcal{X}^{I}$. Secondly, AIFT transforms $\mathcal{X}^{F}$ into the image samples $\bar{\mathcal{X}}^{I}$ via $G^{-}$. Road defects are defected by comparing the given image sample $\mathcal{X}^{I}$ with the transformed result $\bar{\mathcal{X}}^{I}$.

Similarity metric for comparing the two samples $\mathcal{X}^{I}$ and $\bar{\mathcal{X}}^{I}$, is defined as follows, 
\begin{equation}
\begin{split}
\label{eq:sim}
d(\mathcal{X}^{I},\bar{\mathcal{X}}^{I})&=\sum_{i,j}(\bar{x}^{I}_{i,j}\text{log}\frac{\bar{x}^{I}_{i,j}}{m_{i,j}}-x^{I}_{i,j}\text{log}\frac{x^{I}_{i,j}}{m_{i,j}}),
\end{split}
\end{equation}
where $m_{i,j}$ is expectation of ${x}^{I}_{i,j}$ and $\bar{x}^{I}_{i,j}$. Above similarity metric is based on Jeffery divergence, which is a modified KL-divergence to take symmetric property. Euclidean distances such as $l1$-norm and $l2$-normal are not suitable as a similarity metric for images since neighboring values are not considered \cite{rubner2000earth}. Jeffrey divergence is numerically stable, symmetric, and invariant to noise and input scale \cite{puzicha1997non}.

\begin{table}
	\centering
		\caption{Quantitative performance comparison of the detection performance on AIFT using GAPs384 dataset and CFD dataset depending on the loss functions $\mathcal{L}_{\text{re}}$ (Eq \ref{eq:recon_loss}), $\mathcal{L}_{\text{ATCL}}$ (Eq \ref{eq:atcl_loss}), and $\mathcal{L}_{\text{total}}$ (Eq \ref{eq:total_loss}). The bolded figures indicate the best performances on the experiments.}
	\resizebox{\columnwidth}{!}{	
	\begin{tabular}{l c c c  c c c}
		\toprule
		\multirow{2}[4]{*}{Model}  & \multicolumn{3}{c}{GAPs384 dataset \cite{eisenbach2017get}} & \multicolumn{3}{c}{CFD dataset \cite{shi2016automatic}}  \\ \cmidrule(rl){2-7}
		 & AIU  & ODS  & OIS & AIU  & ODS  & OIS       \\ 
		\cmidrule(r){1-7}%f\cmidrule(l){2-7}
		\cmidrule(r){1-7}%f\cmidrule(l){2-7}
		\multicolumn{1}{l}{AIFT$_{\text{re}}$}& 0.052 & 0.181  & 0.201 & 0.152 & 0.562 & 0.572 \\
		\cmidrule(r){1-7}
		\cmidrule(r){1-7}
		\multicolumn{1}{l}{AIFT$_{\text{GAN}}$}& 0.081 & 0.226  & 0.234 & 0.187 & 0.642 & 0.659 \\
		\cmidrule(r){1-7}
		\cmidrule(r){1-7}
		\multicolumn{1}{l}{AIFT$_{\text{total}}$}& \textbf{0.083} & \textbf{0.247}  & \textbf{0.249} & \textbf{0.203} & \textbf{0.701} & \textbf{0.732} \\
		\bottomrule
	\end{tabular}
	}
	%\centering
	\label{tbl:1}
	\vspace{-4ex}
\end{table}

\section{EXPERIMENT}
\label{sec:4}
\subsection{Experiment setting and dataset} 
\label{sec:4:1}
To evaluation the performance of the proposed method on road defect detection, we employ the best F-measure on the dataset for a fixed scale (ODS), the aggregate F-measure on the dataset for the best scale in each image (OIS), and AIU, which is proposed by Yang \etal \cite{yang2019feature}. AIU is computed on the detection and ground truth without non-max suppression (NNS) and thinking operation, defined by, $\frac{1}{N_{t}}\sum_{t}\frac{N^{t}_{pg}}{N^{t}_{p}+N^{t}_{g}-N^{t}_{pg}}$, where $N_{t}$ denotes the total number of thresholds $t\in\{0.01, 0.99\}$ with interval 0.01; for a given $t$, $N^{t}_{pg}$ is the number of pixels of intersected region between the predicted and ground truth crack area; $N^{t}_{p}$ and $N^{t}_{g}$ denote the number of pixels of predicted and ground truth crack region, respectively. The proposed method has been evaluated on four publicly available datasets. The details of the datasets are described as follows.

\textbf{GAPs384 dataset} is German Asphalt Pavement Distress (GAPs) dataset presented by Eisenbach \etal \cite{eisenbach2017get}, and it is constructed to address the issue of comparability in the pavement distress domain by providing a standardized high-quality dataset of large scale. The dataset contains 1,969 gray scaled images for road defects, with various classes for defects fsuch as cracks, potholes, and inlaid patches. The resolution of images is 1,920$\times$1,080.

\textbf{Cracktree200 dataset} \cite{zou2012cracktree} contains 206 road pavement images with 800$\times$600 resolution, which can be categorized to various types of pavement defects. The images on this dataset are captured with some challenging issues such as shadows, occlusions, low contrast, and noise.

\textbf{CRACK500 dataset} is constructed by Yang \etal \cite{yang2019feature}. The dataset is composed of 500 images wity 2,000$\times$1,500, and each image has a pixel-level annotation. The dataset is seperated by training dataset and test dataset. The training dataset consists of 1,896 images, and the test dataset is composed of 1,124 images.

\textbf{CFD dataset} \cite{shi2016automatic} contains 118 images with 480$\times$320 resolution. Each image has pixel-level annotation and captured by Iphone 5 with focus of 4mm aperture of $f/2.4$ and exposure time of 1/135s.

The hyperparameter setting for the best performance is as follows. The epoch size and the batch size are 50 and 64, respectively. The balancing weight for the reconstruction loss $E_{re}$ is 0.1, and the critic iteration is set by 10. The networks are optimized by Adam \etal \cite{kingma2014adam}. The proposed approach has implemented with Pytorch library \footnote{Source codes are publicly available on \url{https://github.com/andreYoo/Adversarial-IFTN.git}}, and the experiments have conducted with GTX Titan XP and 32GB memory.

\begin{figure*}[ht]
	\vspace{-0.2cm}
	\centering
	\includegraphics[width=0.9\textwidth,height=6cm]{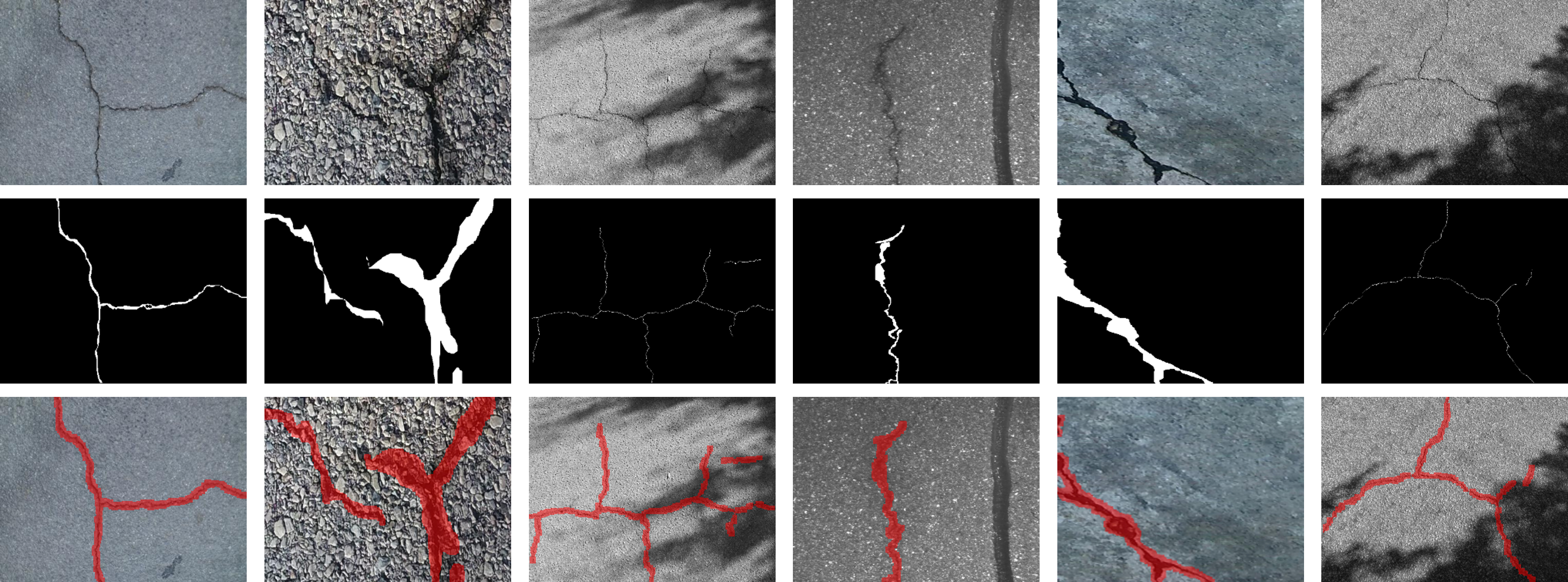}
	\caption{Visualization of the road defect detection results. The images on the first row represent the input images. The second row's images illustrate the ground-truths. The images on the third row denote the detection results for road defects.}
	\label{fig:exp_shots}
	\vspace{-0.2cm}
\end{figure*}

\begin{table*}[ht]
	\centering
	\begin{tabular}{l c c c c c c c c c c c c c c c c c}
		\toprule
		\midrule
		\multirow{2}[4]{*}{Methods} &
		\multirow{2}[4]{*}{S/U} &
		\multicolumn{3}{c}{GAPs384 \cite{eisenbach2017get}} & 
		\multicolumn{3}{c}{Cracktree200 \cite{zou2012cracktree}} & 
		\multicolumn{3}{c}{CRACK500 \cite{yang2019feature}}  & 
		\multicolumn{3}{c}{CFD \cite{shi2016automatic}} &
		\multirow{1}[4]{*}{$\text{FPS(s)}$} \\ 
		\cmidrule(l){3-5} \cmidrule(l){6-8} \cmidrule(l){9-11} \cmidrule(l){12-14} 
		&  & AIU  & ODS  & OIS  & AIU & ODS  & OIS  & AIU  & ODS  & OIS  & AIU & ODS  & OIS &  &\\
		\cmidrule(l){1-16}
		\multicolumn{1}{l}{HED \cite{xie2015holistically}}& S & 0.069 & 0.209 & 0.175 & 0.040 & 0.317 & 0.449 &0.481 & 0.575 & 0.625 &0.154 &0.683 &0.705 & 0.0825\\
		\multicolumn{1}{l}{RCF \cite{liu2017richer}}& S & 0.043 & 0.172  & 0.120 & 0.032  & 0.255 & 0.487 & 0.403 &0.490 & 0.586 &0.105 & 0.542 & 0.607 & \textbf{0.079} \\
		\multicolumn{1}{l}{FCN \cite{long2015fully}}& S & 0.015 & 0.088  & 0.091 & 0.008  & 0.334 & 0.333   &0.379 & 0.513&  0.577 & 0.021 & 0.585 & 0.609 & 0.114\\
		\multicolumn{1}{l}{CrackForest \cite{shi2016automatic}}& U & - & 0.126  & 0.126 & -  & 0.080 & 0.080   & - & 0.199 & 0.199 &-&0.104 & 0.104 & 3.971\\
		\multicolumn{1}{l}{FPHBN \cite{yang2019feature}}& S & 0.081 & 0.220  & 0.231 & 0.041  & 0.517 & 0.579   & \textbf{0.489} & \textbf{0.604} & \textbf{0.635} & 0.173 & 0.683 & 0.705 & 0.237\\
		\multicolumn{1}{l}{AAE \cite{makhzani2015adversarial}}& U & 0.062 & 0.196  & 0.202 & 0.039 & 0.472 & 0.491 & 0.371 & 0.481 & 0.583 & 0.142 & 0.594 & 0.613 & 0.721 \\
		\multicolumn{1}{l}{SVM \cite{zhang2016road}}& S & 0.051 & 0.132  & 0.162 & 0.017 & 0.382 & 0.391 & 0.362 & 0.418 & 0.426 & 0.082 & 0.3R52 & 0.372 & 0.852\\
		\multicolumn{1}{l}{ConvNet \cite{zhang2016road}}& S & 0.079 & 0.203  & 0.211 & 0.037 & 0.472 &  0.499 & 0.431 & 0.591 & 0.609 & 0.152 & 0.579 & 0.677 & 0.921\\
		\midrule
		\multicolumn{1}{l}{AIFT$_{\text{total}}$} & & \textbf{0.083} & \textbf{0.247}  & \textbf{0.249} & \textbf{0.045} & \textbf{0.607} & \textbf{0.642} & 0.478 & 0.549  & 0.561 & \textbf{0.203} & \textbf{0.701} & \textbf{0.732} & 1.1330 \\
		\midrule
		\bottomrule
	\end{tabular}
	\caption{Quantitative performance comparison about road defect detection using GAPs384 \cite{eisenbach2017get}, Cracktree200 \cite{zou2012cracktree}, CRACK500 \cite{yang2019feature}, and CFD \cite{shi2016automatic}. "-" means the results are not provided. The \textbf{bolded} figures indicate that the best performance among them. 'S/U' denotes whether a model focuses on \textit{'supervised'} or  \textit{'unsupervised'} approaches. $\text{FPS}$ indicates the execution speed of each method, and it is computed by averaging the execution speeds about all datasets.}
	\label{tbl:2}
	\vspace{-3ex}
\end{table*}

\subsection{Ablation study}
\label{sec:4:2}
We have conducted an ablation study to observe the effect of the loss function terms on the performance of AIFT. We have trained AIFT using the three loss functions $\mathcal{L}_{\text{re}}$ (Eq \ref{eq:recon_loss}), $\mathcal{L}_{\text{ATCL}}$ (Eq \ref{eq:atcl_loss}), and $\mathcal{L}_{\text{total}}$ (Eq \ref{eq:total_loss}) using GAPs384 dataset and CFD dataset, and observed AIU at every two epochs. The hyperparameter settings applied to train each model, are all same, and only the loss functions are different. Fig \ref{fig:ab_study} shows the AIU trends of AIFTs trained by the three loss functions. Table \ref{tbl:1} contains AIUs, ODSs, and OISs on GAPs384 dataset and CFD dataset. The experimental results show that AIFT trained by the total loss (AIFT$_{\text{total}}$) achieves the best performance on this experiments. As shown in Table \ref{tbl:1}, AIFT$_{\text{total}}$ achieves 0.083 of AIU, 0.247 of OIS, and 0.249 of ODS for GAPs384 dataset. These figures show that AIFT$_{\text{total}}$ can produce approximately 7\% better performance than others. In the experiments using CFD dataset, AIFT$_{\text{total}}$ achieves 0.203 of AIU, 0.701 of OIS, and 0.732 of ODS, and these figure are all higher than that of the others. 

Notably, the overall experimental results demonstrate that the AIFTs trained by adversarial learning, can outperform the AIFT based on the reconstruction setting (AIFT$_{\text{re}}$). Not only AIFT$_{\text{total}}$, but also AIFT$_{\text{ATCL}}$ obtains the improved achievement than AIFT$_{\text{re}}$. The AIU Trends (Fig \ref{fig:ab_study}) also justify that the AIFT learnt by adversarial manners can outperform the AIFT trained by the reconstruction setting. The experimental results justify adversarial learning can improve the robustness of AIFT for detecting road defects.

\subsection{Comparison with existing state-of-the-arts}
\label{sec:4:3}
We have carried out the comparison with existing state-of-the-art methods for the crack detection \cite{xie2015holistically,shi2016automatic,yang2019feature} and the road defect detection \cite{zhang2016road}. For the efficiency of the experiments, only AIFT$_{\text{total}}$ is compared with other methods. Table \ref{tbl:2} contains AIUs, OISs, and ODSs on Cracktree200, GAPs384, Cracktree200, and CFD datasets. AIFT$_{\text{total}}$ has achieved state-of-the-art performance for GAPs384 dataset, Cracktree200 dataset, and CFD dataset. In the experiments using GAPs384 dataset, AIFT$_{\text{total}}$ achieves 0.083 of AIU, 0.247 of ODS, and 0.249 of OIS. These figures show that AIFT$_{\text{total}}$ outperforms than the previous state-of-the-art performance that achieved by FPHBN \cite{yang2019feature}. FPHBN obtains 0.081 of AIU, 0.220 of ODS, and 0.231 of OIS. AIFT$_{\text{total}}$ shows 3\% better performances than FPHBN. The experiments on Cracktree200 dataset and CFD dataset also show that AIFT$_{\text{total}}$ surpasses other methods. AIFT$_{\text{total}}$ produces 0.045 of AIU, 0.607 of ODS, and 0.642 of OIS in the experiments using Cracktree200 dataset. Additionally, AIFT$_{\text{total}}$ achieves 0.203 of AIU, 0.701 of ODS, and 0.732 of OIS on CFD dataset. These figures are 8.8\% and 3\% better than the previous state-of-the-art methods.

However, AIFT$_{\text{total}}$ could not obtain the highest performance on CRACK500 dataset. The state-of-the-art performance on CRACK500 dataset is achieved by FPHBN \cite{yang2019feature}, and it produces 0.489 of AIU, 0.604 of ODS, and 0.635 of OIS, respectively. AIFT$_{\text{total}}$ has 0.478 of AIU, 0.549 of ODS, and 0.561 of OIS. The gaps between FPHBN and AIFT$_{\text{total}}$ are 0.011 on AIU, 0.055 on ODS, and 0.074 on OIS. However, FPHBN exploits a supervised approach, and it needs predetermined pixel-level annotations for road defects. Also, the network architecture applied to their approach is much deeper than Ours. These are the great advantages of detecting road defects.

The overall experiments show that AIFT$_{\text{total}}$ can outperform existing state-of-the-art methods. As shown in Table \ref{tbl:2}, the detection performance of AIFT$_{\text{total}}$ surpasses other unsupervised methods \cite{shi2016automatic,makhzani2015adversarial}. Additionally, AIFT$_{\text{total}}$ achieves outstanding detection performance in detecting defects than others based on supervised learning approaches, even AIFT$_{\text{total}}$ does not need an annotation for road defects in the training step. This may be thought that AIFT$_{\text{total}}$ is enabled to apply various practical situations in which a large-scale and well-annotated dataset can not be used. Consequently, the experimental results demonstrate that AIFT$_{\text{total}}$ can outperform existing state-of-the-art methods.

\section{CONCLUSIONS}
\label{sec:5}
In this paper, we have proposed an unsupervised approach to detecting road defects, based on adversarial image-to-frequency transform. The experimental results demonstrate the proposed approach can detect various patterns of road defects without explicit annotations for road defects in the training step, and it outperforms existing state-of-the-art methods in most of the cases for experiments of road defect detection. 

\addtolength{\textheight}{-12cm}   
\section*{ACKNOWLEDGMENT}
This work was partly supported by the ICT R\&D program of MSIP/IITP. (2014-0-00077, Development of global multi target tracking and event prediction techniques based on real-time large-scale video analysis).

\small
\bibliographystyle{ieeetr}
\bibliography{egbib}

\begin{thebibliography}{10}

\bibitem{zaloshnja2009cost}
E.~Zaloshnja and T.~R. Miller, ``Cost of crashes related to road conditions,
  united states, 2006,'' in {\em Annals of Advances in Automotive
  Medicine/Annual Scientific Conference}, vol.~53, p.~141, Association for the
  Advancement of Automotive Medicine, 2009.

\bibitem{carr2018road}
T.~A. Carr, M.~D. Jenkins, M.~I. Iglesias, T.~Buggy, and G.~Morison, ``Road
  crack detection using a single stage detector based deep neural network,'' in
  {\em 2018 IEEE Workshop on Environmental, Energy, and Structural Monitoring
  Systems (EESMS)}, pp.~1--5, IEEE, 2018.

\bibitem{hadavandsiri2019concrete}
Z.~Hadavandsiri, D.~D. Lichti, A.~Jahraus, and D.~Jarron, ``Concrete
  preliminary damage inspection by classification of terrestrial laser scanner
  point clouds through systematic threshold definition,'' {\em ISPRS
  International Journal of Geo-Information}, vol.~8, no.~12, p.~585, 2019.

\bibitem{acosta1992low}
J.~A. Acosta, J.~L. Figueroa, and R.~L. Mullen, ``Low-cost video image
  processing system for evaluating pavement surface distress,'' {\em
  Transportation research record}, no.~1348, 1992.

\bibitem{bray2006neural}
J.~Bray, B.~Verma, X.~Li, and W.~He, ``A neural network based technique for
  automatic classification of road cracks,'' in {\em The 2006 IEEE
  International Joint Conference on Neural Network Proceedings}, pp.~907--912,
  IEEE, 2006.

\bibitem{chambon2010road}
S.~Chambon, C.~Gourraud, J.~M. Moliard, and P.~Nicolle, ``Road crack extraction
  with adapted filtering and markov model-based segmentation: introduction and
  validation,'' 2010.

\bibitem{deutschl2004defect}
E.~Deutschl, C.~Gasser, A.~Niel, and J.~Werschonig, ``Defect detection on rail
  surfaces by a vision based system,'' in {\em IEEE Intelligent Vehicles
  Symposium, 2004}, pp.~507--511, IEEE, 2004.

\bibitem{sun2009automated}
Y.~Sun, E.~Salari, and E.~Chou, ``Automated pavement distress detection using
  advanced image processing techniques,'' in {\em 2009 IEEE International
  Conference on Electro/Information Technology}, pp.~373--377, IEEE, 2009.

\bibitem{baygin2015new}
M.~Baygin and M.~Karakose, ``A new image stitching approach for resolution
  enhancement in camera arrays,'' in {\em 2015 9th International Conference on
  Electrical and Electronics Engineering (ELECO)}, pp.~1186--1190, IEEE, 2015.

\bibitem{pauly2017deeper}
L.~Pauly, D.~Hogg, R.~Fuentes, and H.~Peel, ``Deeper networks for pavement
  crack detection,'' in {\em Proceedings of the 34th ISARC}, pp.~479--485,
  IAARC, 2017.

\bibitem{fan2019road}
R.~Fan, M.~J. Bocus, Y.~Zhu, J.~Jiao, L.~Wang, F.~Ma, S.~Cheng, and M.~Liu,
  ``Road crack detection using deep convolutional neural network and adaptive
  thresholding,'' in {\em 2019 IEEE Intelligent Vehicles Symposium (IV)},
  pp.~474--479, IEEE, 2019.

\bibitem{abdel2006pca}
I.~Abdel-Qader, S.~Pashaie-Rad, O.~Abudayyeh, and S.~Yehia, ``Pca-based
  algorithm for unsupervised bridge crack detection,'' {\em Advances in
  Engineering Software}, vol.~37, no.~12, pp.~771--778, 2006.

\bibitem{oliveira2012automatic}
H.~Oliveira and P.~L. Correia, ``Automatic road crack detection and
  characterization,'' {\em IEEE Transactions on Intelligent Transportation
  Systems}, vol.~14, no.~1, pp.~155--168, 2012.

\bibitem{mujeeb2019one}
A.~Mujeeb, W.~Dai, M.~Erdt, and A.~Sourin, ``One class based feature learning
  approach for defect detection using deep autoencoders,'' {\em Advanced
  Engineering Informatics}, vol.~42, p.~100933, 2019.

\bibitem{kang2018deep}
G.~Kang, S.~Gao, L.~Yu, and D.~Zhang, ``Deep architecture for high-speed
  railway insulator surface defect detection: Denoising autoencoder with
  multitask learning,'' {\em IEEE Transactions on Instrumentation and
  Measurement}, 2018.

\bibitem{vincent2010stacked}
P.~Vincent, H.~Larochelle, I.~Lajoie, Y.~Bengio, and P.-A. Manzagol, ``Stacked
  denoising autoencoders: Learning useful representations in a deep network
  with a local denoising criterion,'' {\em Journal of machine learning
  research}, vol.~11, no.~Dec, pp.~3371--3408, 2010.

\bibitem{perera2019ocgan}
P.~Perera, R.~Nallapati, and B.~Xiang, ``Ocgan: One-class novelty detection
  using gans with constrained latent representations,'' in {\em Proceedings of
  the IEEE Conference on Computer Vision and Pattern Recognition},
  pp.~2898--2906, 2019.

\bibitem{pidhorskyi2018generative}
S.~Pidhorskyi, R.~Almohsen, and G.~Doretto, ``Generative probabilistic novelty
  detection with adversarial autoencoders,'' in {\em Advances in Neural
  Information Processing Systems}, pp.~6822--6833, 2018.

\bibitem{zhu2017unpaired}
J.-Y. Zhu, T.~Park, P.~Isola, and A.~A. Efros, ``Unpaired image-to-image
  translation using cycle-consistent adversarial networks,'' in {\em
  Proceedings of the IEEE international conference on computer vision},
  pp.~2223--2232, 2017.

\bibitem{ying2019x2ct}
X.~Ying, H.~Guo, K.~Ma, J.~Wu, Z.~Weng, and Y.~Zheng, ``X2ct-gan:
  Reconstructing ct from biplanar x-rays with generative adversarial
  networks,'' in {\em Proceedings of the IEEE Conference on Computer Vision and
  Pattern Recognition}, pp.~10619--10628, 2019.

\bibitem{bai2018finding}
Y.~Bai, Y.~Zhang, M.~Ding, and B.~Ghanem, ``Finding tiny faces in the wild with
  generative adversarial network,'' in {\em Proceedings of the IEEE Conference
  on Computer Vision and Pattern Recognition}, pp.~21--30, 2018.

\bibitem{liu2018future}
W.~Liu, W.~Luo, D.~Lian, and S.~Gao, ``Future frame prediction for anomaly
  detection--a new baseline,'' in {\em Proceedings of the IEEE Conference on
  Computer Vision and Pattern Recognition}, pp.~6536--6545, 2018.

\bibitem{sabokrou2018deep}
M.~Sabokrou, M.~Fayyaz, M.~Fathy, Z.~Moayed, and R.~Klette, ``Deep-anomaly:
  Fully convolutional neural network for fast anomaly detection in crowded
  scenes,'' {\em Computer Vision and Image Understanding}, vol.~172,
  pp.~88--97, 2018.

\bibitem{rubner2000earth}
Y.~Rubner, C.~Tomasi, and L.~J. Guibas, ``The earth mover's distance as a
  metric for image retrieval,'' {\em International journal of computer vision},
  vol.~40, no.~2, pp.~99--121, 2000.

\bibitem{puzicha1997non}
J.~Puzicha, T.~Hofmann, and J.~M. Buhmann, ``Non-parametric similarity measures
  for unsupervised texture segmentation and image retrieval,'' in {\em
  Proceedings of IEEE Computer Society Conference on Computer Vision and
  Pattern Recognition}, pp.~267--272, IEEE, 1997.

\bibitem{eisenbach2017get}
M.~Eisenbach, R.~Stricker, D.~Seichter, K.~Amende, K.~Debes, M.~Sesselmann,
  D.~Ebersbach, U.~Stoeckert, and H.-M. Gross, ``How to get pavement distress
  detection ready for deep learning? a systematic approach,'' in {\em 2017
  international joint conference on neural networks (IJCNN)}, pp.~2039--2047,
  IEEE, 2017.

\bibitem{shi2016automatic}
Y.~Shi, L.~Cui, Z.~Qi, F.~Meng, and Z.~Chen, ``Automatic road crack detection
  using random structured forests,'' {\em IEEE Transactions on Intelligent
  Transportation Systems}, vol.~17, no.~12, pp.~3434--3445, 2016.

\bibitem{yang2019feature}
F.~Yang, L.~Zhang, S.~Yu, D.~Prokhorov, X.~Mei, and H.~Ling, ``Feature pyramid
  and hierarchical boosting network for pavement crack detection,'' {\em IEEE
  Transactions on Intelligent Transportation Systems}, 2019.

\bibitem{zou2012cracktree}
Q.~Zou, Y.~Cao, Q.~Li, Q.~Mao, and S.~Wang, ``Cracktree: Automatic crack
  detection from pavement images,'' {\em Pattern Recognition Letters}, vol.~33,
  no.~3, pp.~227--238, 2012.

\bibitem{kingma2014adam}
D.~P. Kingma and J.~Ba, ``Adam: A method for stochastic optimization,'' {\em
  arXiv preprint arXiv:1412.6980}, 2014.

\bibitem{xie2015holistically}
S.~Xie and Z.~Tu, ``Holistically-nested edge detection,'' in {\em Proceedings
  of the IEEE international conference on computer vision}, pp.~1395--1403,
  2015.

\bibitem{liu2017richer}
Y.~Liu, M.-M. Cheng, X.~Hu, K.~Wang, and X.~Bai, ``Richer convolutional
  features for edge detection,'' in {\em Proceedings of the IEEE conference on
  computer vision and pattern recognition}, pp.~3000--3009, 2017.

\bibitem{long2015fully}
J.~Long, E.~Shelhamer, and T.~Darrell, ``Fully convolutional networks for
  semantic segmentation,'' in {\em Proceedings of the IEEE conference on
  computer vision and pattern recognition}, pp.~3431--3440, 2015.

\bibitem{makhzani2015adversarial}
A.~Makhzani, J.~Shlens, N.~Jaitly, I.~Goodfellow, and B.~Frey, ``Adversarial
  autoencoders,'' {\em arXiv preprint arXiv:1511.05644}, 2015.

\bibitem{zhang2016road}
L.~Zhang, F.~Yang, Y.~D. Zhang, and Y.~J. Zhu, ``Road crack detection using
  deep convolutional neural network,'' in {\em 2016 IEEE international
  conference on image processing (ICIP)}, pp.~3708--3712, IEEE, 2016.

\end{thebibliography}
\end{document}